\documentclass[review]{elsarticle}

\usepackage{lineno,hyperref}
\usepackage{verbatim}
\usepackage{placeins}
\usepackage{graphicx}
\usepackage{subcaption}
\modulolinenumbers[5]
\usepackage{amsmath}
\usepackage{amsmath}
\usepackage{amsfonts}
\usepackage{amssymb}

\journal{Astronomy and Computing}

\biboptions{authoryear}

\usepackage[left=2cm,top=2cm,right=2cm,bottom=2cm,footskip=1cm]{geometry}
\usepackage{multirow}
\usepackage{diagbox}
\begin{document}

\begin{frontmatter}

\title{SBAF: A New Activation Function for Artificial Neural Net based Habitability Classification}

\author{Snehanshu Saha}
\address{PES University South Campus}
\author{Archana Mathur}
\address{Indian Statistical Institute}
\author{Kakoli Bora, Surbhi Agrawal}
\address{PES University South Campus}
\author{Suryoday Basak}
\address{University of Texas at Arlington}






\begin{abstract}
We explore the efficacy of using a novel activation function in Artificial Neural Networks (ANN) in characterizing exoplanets into different classes. We call this Saha-Bora Activation Function (SBAF) as the motivation is derived from long standing understanding of using advanced calculus in modeling habitability score of Exoplanets. The function is demonstrated to possess nice analytical properties and doesn't seem to suffer from local oscillation problems. The manuscript presents the analytical properties of the activation function and the architecture implemented on the function.
\end{abstract}


\end{frontmatter}


\section{Introduction}

For hundreds of years, astronomers and philosophers have considered the possibility that the Earth is a very rare case of a planet as it harbors life. This was partly due to the fact that after the initial missions exploring our neighbors Mars and Venus, no traces of life were found. However, over the past two decades, discoveries of exoplanets have poured in by the hundreds and the rate at which exoplanets are being discovered is increasing. The inference from this is that planets around stars are a rule rather than an exception with the actual number of planets exceeding the number of stars in our galaxy by orders of magnitude. In order to find interesting samples from the massive ongoing growth in the data, a sophisticated pipeline may be developed which can quickly and efficiently classify exoplanets based on habitability classes.

The process of discovery of exoplanets is rather complex, \citep{Bains2016}, as the size of exoplanets is small compared to other types of stellar objects such as stars, galaxies, quasars, etc. which can be discovered with greater ease. A very careful analysis of stellar signals is required to detect planetary samples. Some of the methods of detecting exoplanets include radial velocity based detections, gravitational lensing, etc. Imaging-based methods of discovery of exoplanets are not well developed yet and are at a rather controversial stage but could be more effective in exoplanet discovery with improvements. The data collected is imperfect and sometimes difficult to analyze with certainty. Given the rapid technological improvements and the accumulation of a large amount of data, it is pertinent to explore advanced methods of data analysis to rapidly classify planets into appropriate categories based on the physical characteristics.

There exist different approaches to solving the habitability problem. Explicit score computation, \citep{CDHPF2016} giving rise to metrics is one way of addressing the issue. However, habitability is too complex a problem to be equated with Earth-similarity alone \citep{1804.11176}. Therefore, model based evaluations \citep{1803.04644} need to be synthesized with feature based classification \citep{Insight}.

Existing work on characterizing exoplanets are based on assigning habitability scores to each planet which allows for a quantitative comparison to Earth. The Earth Similarity Index, Biological Complexity Index and Planetary Habitability Index  are distance-based metrics which gauge the similarity of a planet to that of Earth; the Cobb-Douglas Habitability Score (CDHS), \cite{CDHPF2016} makes use of econometric modeling to find the similarity of a planet to Earth. Recently, a collaborative effort between Google and NASA resulted in the discovery of two exoplanets. In Saha et.al. \cite{Potential}, an advanced tree-based classifier, Gradient Boosted Decision Tree was used to classify Proxima b and planets in the TRAPPIST-1 system. The accuracies were nearly perfect, giving us the basis of exploring other machine classifiers for the task.

Remainder of the paper is organized as follows. A novel activation function to train an artificial neural network (ANN) is introduced. We discuss the theoretical nuances of such a function. In the next section, the back propagation mechanism with the relevant architecture is described paving the foundation for ANN based classification of exoplanets. We conclude by discussing the efficacy of the proposed method.

\section{Saha-Bora Activation Function (SBAF) for a Neural Network}
Neural networks \citep{neural}, commonly known as Artificial Neural network(ANN), is a system of interconnected units organized in layers, which processes information signals by responding dynamically to inputs. Layers of the network are oriented in such a way that inputs are fed at input layer and output layer receives output after being processed at neurons of one or more hidden layers. Hidden layers consist of computing neurons that are connected to input and output layers through a system of weighted connections. The network has ability to learn from input patterns, whereby with every input fed to the network, weights are updated in such a way that the error between the desired and observed output is minimum. Hidden layers are equipped with a special function called activation function \citep{ELFWING2018}, \citep{sbaf} to trigger neurons to process and propagate outputs across the network. \\
A special class of ANN called Back propagation \citep{back} deals with computing the error between observed and desired output and later feeds this error back to the network with each cycle or 'epoch'. The weights are updated correspondingly and learning or training of the network is performed till the error is minimized. \\
Activation function acts as a functional mapping between inputs and outputs. It allows the network to learn and model complex dataset like audio, video and text. Most popular activation functions are Sigmoid, hyperbolic tangent and Relu.
\par The activation function is as follows:
\begin{equation}
\begin{split}
y &= \frac{1}{1 + kx^{\alpha}(1-x)^{1-\alpha}}; \\
\Rightarrow \textrm{ln}y &= \textrm{ln}1 - \textrm{ln}(1 + kx^{\alpha}(1-x)^{1-\alpha})\\
 &= - \textrm{ln}(1 + kx^{\alpha}(1-x)^{1-\alpha})\\
\Rightarrow \frac{1}{y}\frac{dy}{dx} &= - \frac{1}{(1 + kx^{\alpha}(1-x)^{1-\alpha})} \cdot \Big[k\alpha x^{\alpha -1} (1-x)^{1-\alpha} - kx^{\alpha}(1-\alpha)(1-x)^{1 - \alpha -1} \Big]\\
&= - \frac{k}{(1 + kx^{\alpha}(1-x)^{1-\alpha})} \cdot \Big[\alpha x^{\alpha -1} (1-x)^{1-\alpha} - (1-\alpha)x^{\alpha}(1-x)^{-\alpha} \Big]\\
\Rightarrow \frac{dy}{dx} &= y \Bigg[ \frac{\alpha}{x} - (1 - \alpha)\frac{1}{1-x} \Bigg]kx^{\alpha}(1-x)^{1-\alpha}\\
&= y \Bigg[\frac{\alpha (1-x) - (1-\alpha)x}{x(1-x)} \Bigg]kx^{\alpha}(1-x)^{1-\alpha}\\
&= y^{2} \Bigg[\frac{\alpha - x}{x(1-x)} \Bigg]kx^{\alpha}(1-x)^{1-\alpha}\\
\end{split}
\label{eq:acti_func_1}
\end{equation}

From the definition of the function, we have:
\begin{equation}
\begin{split}
y &= \frac{1}{1 + kx^{\alpha}(1-x)^{1-\alpha}}\\
\Rightarrow kx^{\alpha}(1-x)^{1-\alpha} &= \frac{1-y}{y}
\end{split}
\label{eq:acti_func_2}
\end{equation}

Substituting Equation \ref{eq:acti_func_2} in \ref{eq:acti_func_1},

\begin{equation}
\begin{split}
\frac{dy}{dx} &= y^{2} \cdot \frac{\alpha-x}{x(1-x)} \cdot \frac{1-y}{y}\\
&= \frac{y(1-y)}{x(1-x)}\cdot(\alpha-x)
\end{split}
\label{eq:acti_func_3}
\end{equation}
\textbf{Remark:} $x$ is the lineaar combination of surface temperature, called as input to the NN, and weights (normalized between $0$ and $1$) and $1-x$ is the complement of that, together explaining the perfect discrimination between habitability classes as explained in TSI \citep{Insight}. The motivation of SBAF is derived from this fact of TSI. Using $kx^{\alpha}(1-x)^{1-\alpha}$ shall maximize the width of the two separating hyperplanes in the SVM used in TSI (See the proof below) as the kernel has a global maxima when $0 \leq \alpha \leq 1$. This is equivalent to the CDHS formulation when CD-HPF is written as $ y = kx^{\alpha}(1-x)^{\beta}$ where $\alpha+\beta=1, 0 \leq \alpha \leq 1, 0 \leq \beta \leq 1$, $k$ is suitably assumed to be $1$ (CRS condition), and the representation ensures global maxima (maximum width of the separating hyperplanes) under such constraints, \cite{CDHPF2016}, \cite{Potential}. The new activation function to be used for training a neural network for habitability classification boasts of an optima. Evidently, from the graphical simulations below, we observe less flattening of the function and therefore the formulation should be able to tackle local oscillations more easily as compared to the more generally used sigmoid function. Moreover, since $0 \leq \alpha \leq 1, 0 \leq x \leq 1, 0 \leq 1-x \leq 1$, the variable term in the denominator of SBAF, $kx^{\alpha}(1-x)^{1-\alpha}$ may be approximated to a first order polynomial. This may help us in circumventing expensive floating point operations without compromising the precision.
\begin{figure}[htbp!]
\begin{center}
\includegraphics[width=0.6\columnwidth]{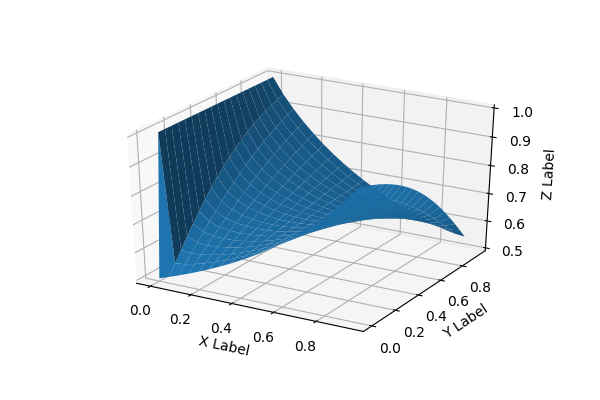}
\caption{Surface Plot of SBAF}
\label{fig:sbaf1}
\end{center}
\end{figure}

\begin{figure}[htbp!]
\begin{center}
\includegraphics[width=0.5\columnwidth]{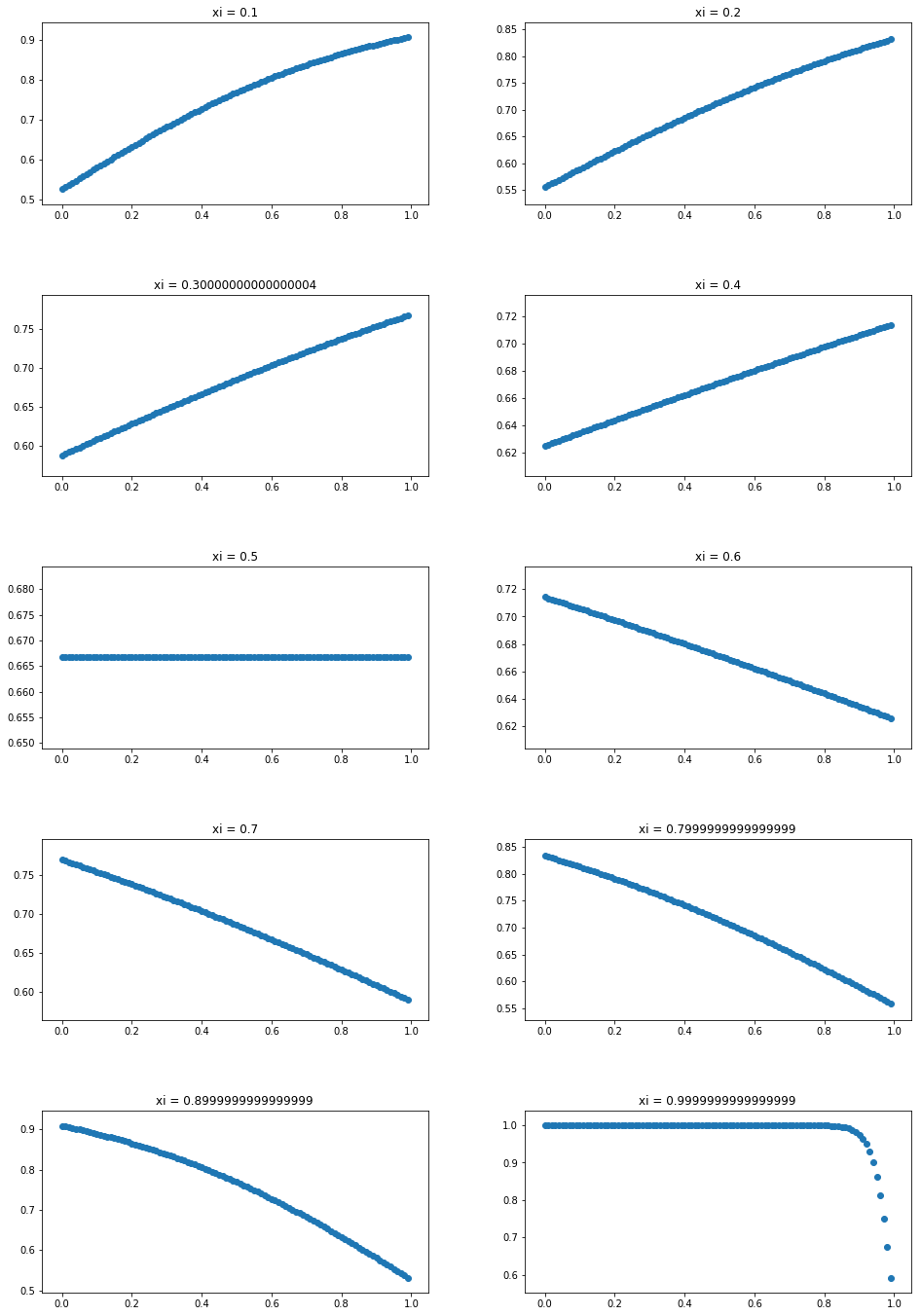}
\caption{}
\label{fig:sbaf2}
\end{center}
\end{figure}

\begin{figure}[htbp!]
\begin{center}
\includegraphics[width=0.5\columnwidth]{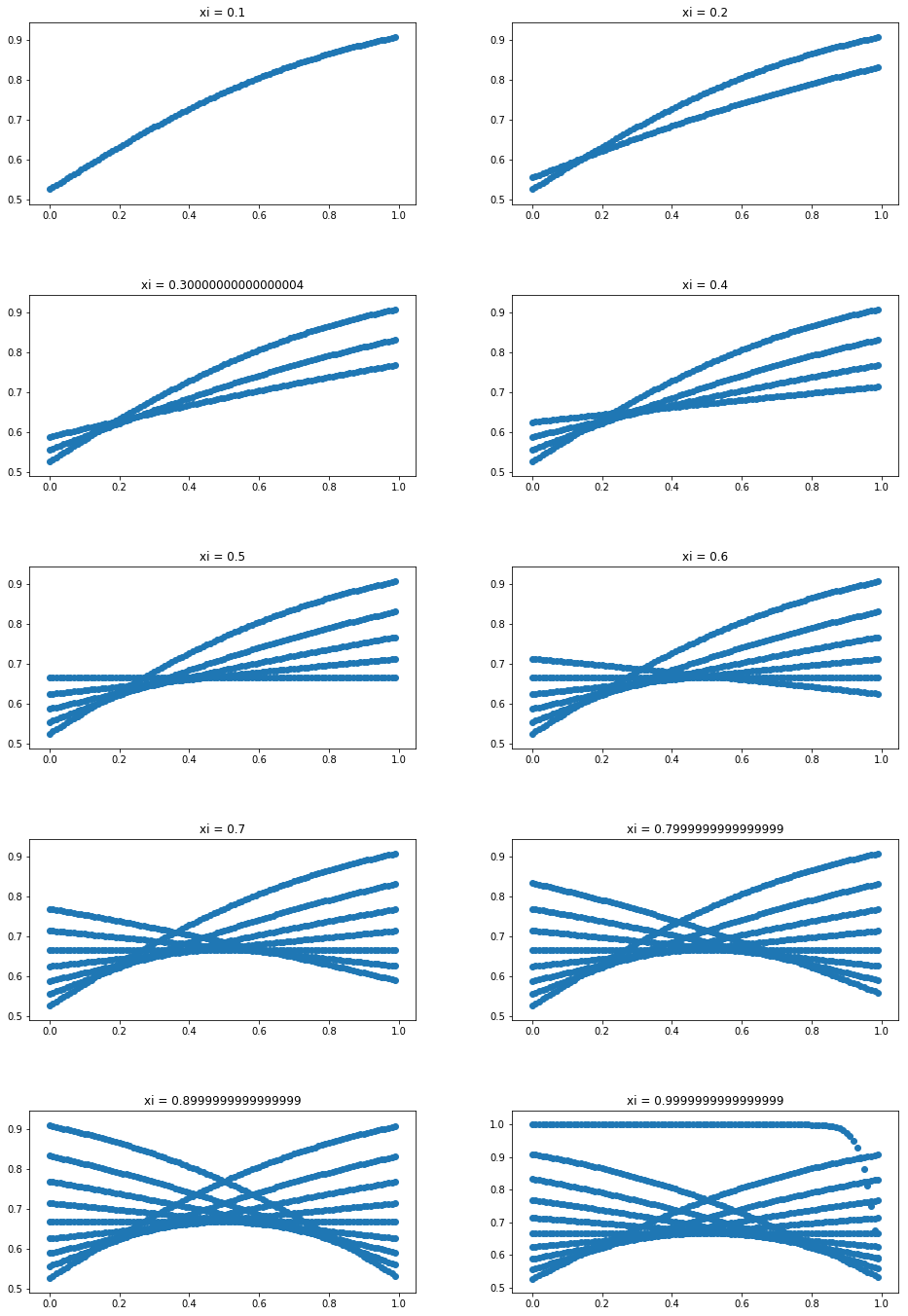}
\caption{}
\label{fig:sbaf3}
\end{center}
\end{figure}
\pagebreak
\subsection{Existence of Optima: Second order Differentiation of SBAF for Neural Network}

From Equation \ref{eq:acti_func_3} ,

\begin{equation}
\begin{split}
\frac{dy}{dx} &=\frac{y(1-y)}{x(1-x)}\cdot(\alpha-x)
\end{split}
\nonumber
\end{equation}
Therefore, \\
\begin{equation}
\begin{split}
\frac{d^{2}y}{dx^{2}} &= \frac{x(1-x)\frac{d[y(1-y)\cdot(\alpha-x)]}{dx}-[y(1-y)\cdot(\alpha-x)]\frac{d(x(1-x))}{dx}}{(x(1-x))^{2}} \\
&= \frac{x(1-x)\frac{d[(y-y^{2})\cdot(\alpha-x)]}{dx}-[(y-y^{2})\cdot(\alpha-x)]\frac{d(x-x^{2})}{dx}}{(x(1-x))^{2}} \\
&= \frac{x(1-x)[(\alpha-x)\frac{d(y-y^{2})}{dx} +(y-y^{2})\cdot \frac{d(\alpha-x)}{dx}] - [(y-y^{2})\cdot(\alpha-x)]\cdot(1-2x)}{(x(1-x))^{2}} \\
&= \frac{x(1-x)[(\alpha-x)\cdot(\frac{dy}{dx} - 2y\frac{dy}{dx}) + (y-y^{2})(-1)]-[(y-y^{2})\cdot(\alpha-x)]\cdot(1-2x)}{(x(1-x))^{2}}\\
\end{split}
\nonumber
\end{equation}

\begin{equation}
\begin{split}
\Rightarrow \frac{d^{2}y}{dx^{2}} &= \frac{x(1-x)[(\alpha-x)\cdot(1-2y)\cdot\frac{dy}{dx} + y(y-1)] + y(y-1)\cdot(\alpha-x)\cdot(1-2x)}{(x(1-x))^{2}}\\
\end{split}
\label{eq:acti_func_5}
\end{equation}

Now, substituting \ref{eq:acti_func_1} in \ref{eq:acti_func_5} we get, \\
\begin{equation}
\begin{split}
\Rightarrow \frac{d^{2}y}{dx^{2}} &= \frac{x(1-x)(\alpha-x)\cdot(1-2y)\cdot\Big[\frac{y(1-y)}{x(1-x)}\cdot(\alpha-x)\Big] + x(1-x)\cdot y(y-1) + y(y-1)\cdot(\alpha-x)\cdot(1-2x)}{(x(1-x))^{2}}\\
&= \frac{y(y-1)[x(1-x)+(\alpha-x)\cdot (1-2x) + (\alpha-x)^{2}\cdot (1-2y)]}{(x(1-x))^{2}} \\
\end{split}
\nonumber
\end{equation}
when $\alpha = x$, \\
\begin{equation}
\begin{split}
\Rightarrow \frac{d^{2}y}{dx^{2}} &= \frac{x(1-x)\cdot y(y-1)}{(x(1-x))^{2}} \\
&= \frac{y(y-1)}{x(1-x)} \\
\end{split}
\nonumber
\end{equation}
Clearly, the first derivative vanishes when $\alpha=x$, the derivative is positive when $\alpha > x$ and is negative when $\alpha < x$ (implying range of values for $\alpha$ so that the function becomes increasing or decreasing, please see Eq. (3)). We need to determine the sign of the second derivative when $\alpha=x$ to ascertain the condition of maxima (corresponding to maximum width of the separating hyperplane ensuring optimal discrimination between habitability classes). Assuming $ 0< x <1$, the condition of optimality, $0 \leq \alpha \leq 1$, $y$ by construction lies between $(0,1)$. Hence, $\frac{d^{2}y}{dx^{2}} < 0$ ensuring maxima of $y$.

\section{Backpropagation with SBAF}
The basic structure of the neural network consists of
input layer, hidden layer and output layer. Let us assume the nodes at input layer are $i_{1}$, $i_{2}$ , at hidden layer $h_{1}$, $h_{2}$ and at output layer $o_{1}$, $o_{2}$.  \subsection{Basic Structure}
\includegraphics[scale=0.5]{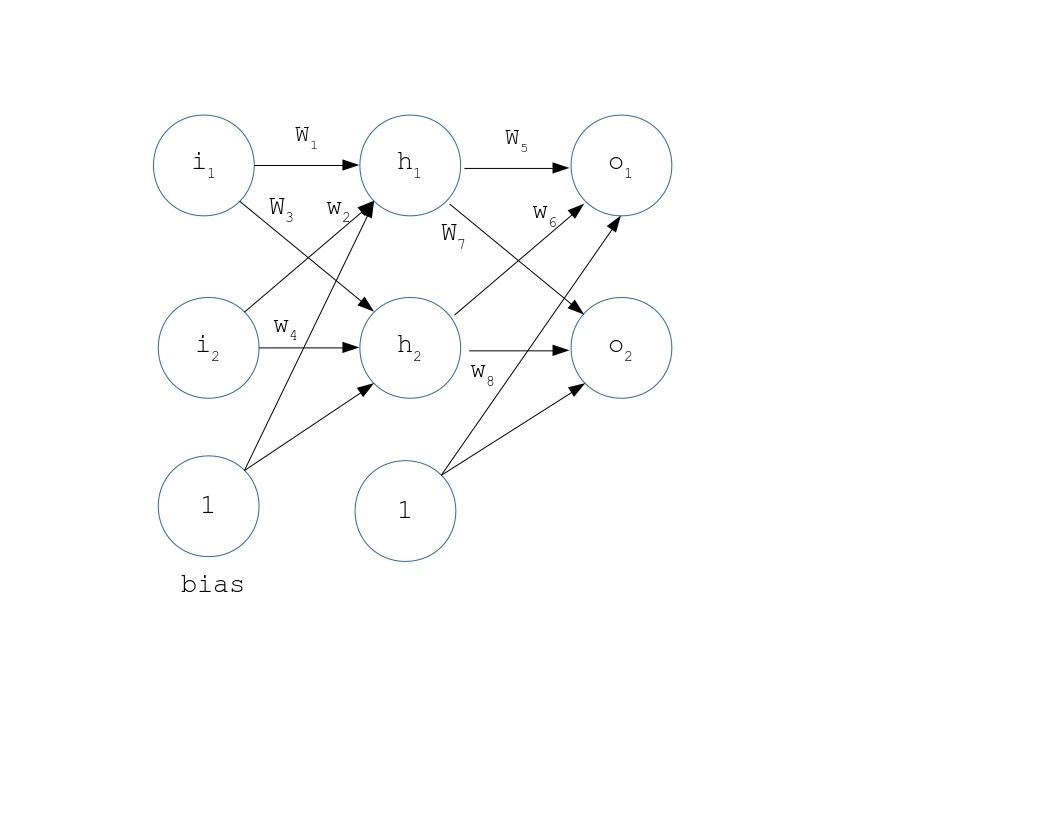} 

\textbf{Goal:} to optimize the weights so that the network can learn how to map from inputs to outputs.
\subsection{The Forward Pass}
Calculate the total input for $h_1$.
$$h1_{net} = w_1 \cdot i_1 + w_2 \cdot i_2 + b_1$$
$$h2_{net} = w_3 \cdot i_1 + w_4 \cdot i_2 + b_1$$
Use SBAF to calculate the output for $h_1$, $y = \frac{1}{1 + kx^\alpha  (1-x)^{1-\alpha}}$.
$$h1_{out} = \frac{1}{1 + k(h1_{net})^\alpha  (1-h1_{net})^{1-\alpha}}$$
$$h2_{out} = \frac{1}{1 + k(h2_{net})^\alpha  (1-h2_{net})^{1-\alpha}}$$
Repeat the process for output layer neuron.
$$o1_{net} = w_5 \cdot h1_{out} + w_6 \cdot h2_{out} + b_2$$
$$o2_{net} = w_7 \cdot h1_{out} + w_8 \cdot h2_{out} + b_2$$
The outputs are
$$o1_{out} = \frac{1}{1 + k(o1_{net})^\alpha  (1-o1_{net})^{1-\alpha}}$$
$$o2_{out} = \frac{1}{1 + k(o2_{net})^\alpha  (1-o2_{net})^{1-\alpha}}$$
Calculating the errors,
$$\mathrm{Error} = \mathrm{Error}_{o1} + \mathrm{Error}_{o2}$$
$$\mathrm{Error}_{o1} = \frac{1}{2}\left( o1_{target} - o1_{out} \right)^2$$
$$\mathrm{Error}_{o2} = \frac{1}{2}\left( o2_{target} - o2_{out} \right)^2$$
\subsection{The Backward Pass}
Update the weights so that the actual output is closer to target output, thereby minimizing the error.
\subsubsection{Output Layer}
Consider $w_5$: let's find the gradient wrt $w_5$, i.e., $\frac{\partial E_{total}}{\partial w_5}$. \\
\includegraphics[scale=0.5]{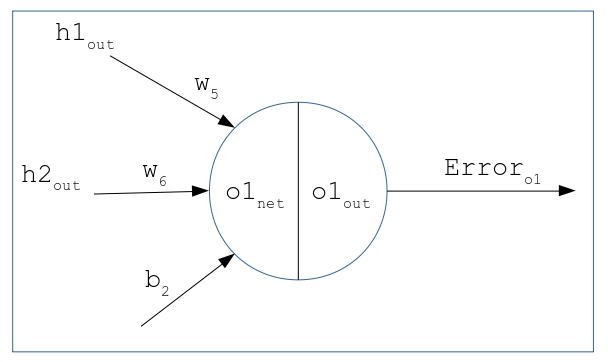} 
\[
\boxed{\frac{\partial E_T}{\partial w_5} = \frac{\partial E_T}{\partial o1_{out}}\cdot \frac{\partial o1_{out}}{\partial o1_{net}} \cdot \frac{\partial o1_{net}}{\partial w_5}}
\]
Calculate each component on the RHS one by one:
\begin{equation}
\boxed{
	\begin{aligned} 
		\partial E_T &= E_{o1} + E_{o2} \\ 
		E_T &= \frac{1}{2} \left( o1_{target} - o1_{out} \right)^2 + \frac{1}{2}\left( o2_{target} - o2_{out} \right)^2 \\ 
		\frac{\partial E}{\partial o1_{out}} &=  2 \cdot \frac{1}{2} \left( o1_{target} - o1_{out} \right) \cdot (-1) + 0 \\
		\frac{\partial E}{\partial o1_{out}} &= -\left( o1_{target} - o1_{out} \right)
	\end{aligned}
} \qquad
\end{equation}
Using the SBAF
\begin{equation}
\boxed{
	\begin{aligned}
		o1_{out} &= \frac{1}{1 + k(o1_{net})^\alpha (1-o1_{net})^{1-\alpha}} \\
		\frac{\partial o1_{out}}{\partial o1_{net}} &= \frac{o1_{out}(1 - o1_{out})}{o1_{net}(1-o1_{net})} \cdot (\alpha - o1_{net})
	\end{aligned}
} \qquad
\end{equation}
Finally,
\begin{equation}
\boxed{
	\begin{aligned}
		o1_{net} &= w_5 \cdot h1_{out} + w_6 \cdot h2_{out} + b_2 \\
		\frac{\partial o1_{net}}{\partial w_5} &= h1_{out}
	\end{aligned}
} \qquad
\end{equation}
Putting $(1)$ and $(2)$ and $(3)$ together in $\frac{\partial E_T}{\partial w_5}$,
\begin{equation*}
\frac{\partial E_T}{\partial w_5} = -\left( o1_{target} - o1_{out} \right) \cdot \frac{o1_{out}(1 - o1_{out})}{o1_{net}(1-o1_{net})} \cdot (\alpha - o1_{net}) \cdot h1_{out}
\end{equation*}
\begin{equation}
w_5^{new} = w_5 - \eta \cdot \frac{\partial E_T}{\partial w_5}
\end{equation}
where $\eta$ is the learning rate. \\
Likewise,
\begin{equation*}
\frac{\partial E_T}{\partial w_6} = -\left( o1_{target} - o1_{out} \right) \cdot \frac{o1_{out}(1 - o1_{out})}{o1_{net}(1-o1_{net})} \cdot (\alpha - o1_{net}) \cdot h2_{out} \\
\end{equation*}
\begin{equation*}
\frac{\partial E_T}{\partial w_7} = -\left( o2_{target} - o2_{out} \right) \cdot \frac{o2_{out}(1 - o2_{out})}{o2_{net}(1-o2_{net})} \cdot (\alpha - o2_{net}) \cdot h1_{out}
\end{equation*}
\begin{equation*}
\frac{\partial E_T}{\partial w_8} = -\left( o2_{target} - o2_{out} \right) \cdot \frac{o2_{out}(1 - o2_{out})}{o2_{net}(1-o2_{net})} \cdot (\alpha - o2_{net}) \cdot h2_{out}
\end{equation*}

\subsubsection{Hidden Layer}
Consider $w_1$ \\
\includegraphics[scale=0.5]{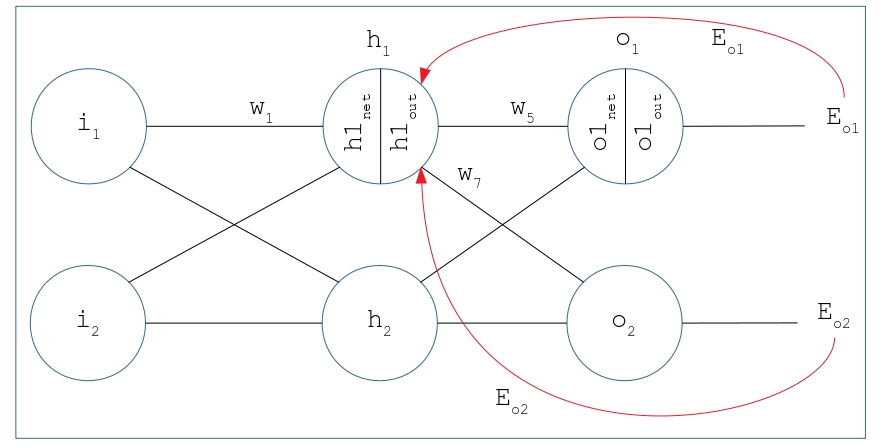}  \\
We need to find $\frac{\partial E_T}{\partial w_1}$. \\
Apparently, \[\frac{\partial E_T}{\partial w_1} = \frac{\partial E_{o1}}{\partial w_1} + \frac{\partial E_{o2}}{\partial w_1}\]
The chain rule says,
\setcounter{equation}{0}
\begin{equation}
\frac{\partial E_{o1}}{\partial w_1} = \frac{\partial E_{o1}}{\partial o1_{out}} \cdot \frac{\partial o1_{out}}{\partial o1_{net}} \cdot \frac{\partial o1_{net}}{\partial h1_{out}} \cdot \frac{\partial h1_{out}}{\partial h1_{net}} \cdot \frac{\partial h1_{net}}{\partial w_1}
\end{equation}
\begin{equation}
\frac{\partial E_{o2}}{\partial w_1} = \frac{\partial E_{o2}}{\partial o2_{out}} \cdot \frac{\partial o2_{out}}{\partial o2_{net}} \cdot \frac{\partial o2_{net}}{\partial h1_{out}} \cdot \frac{\partial h1_{out}}{\partial h1_{net}} \cdot \frac{\partial h1_{net}}{\partial w_1}
\end{equation}
Computing all the components of equation $(1)$,
\begin{align*}
\frac{\partial E_{o1}}{\partial o1_{out}} &= -(o1_{target} - o1_{out}) \\
\frac{\partial o1_{out}}{\partial o1_{net}} &= \frac{o1_{out}(1-o1_{out})}{o1_{net}(1-o1_{net})} \cdot (\alpha - o1_{net}) \\
\frac{\partial o1_{net}}{\partial h1_{out}} &= w_5 \left(\because o1_{net} = w_5 \cdot h1_{out} + w_6 \cdot h2_{out} + b_2 \text{ and } \frac{\partial o1_{net}}{\partial h1_{out}} = w_5 \right) \\
\frac{\partial h1_{out}}{\partial h1_{net}} &= \frac{h1_{out}(1-h1_{out})}{h1_{net}(1-h1_{net})} \cdot (\alpha - h1_{net}) \\
\frac{\partial h1_{net}}{\partial w_1} &= i_1 \left(\because h1_{net} = w_1 i_1 + w_2 i_2 + b_1 \text{ and } \frac{\partial h1_{net}}{\partial w_1} = i_1 + 0 \right)
\end{align*}
Similarly, computing all the components of $(2)$,
\begin{align*}
\frac{\partial E_{o2}}{\partial o2_{out}} &= -(o2_{target} - o2_{out}) \\
\frac{\partial o2_{out}}{\partial o2_{net}} &= \frac{o2_{out}(1-o2_{out})}{o2_{net}(1-o2_{net})} \cdot (\alpha - o2_{net}) \\
\frac{\partial o2_{net}}{\partial h1_{out}} &= w_7 \left(\because o2_{net} = w_7 h1_{out} + w_8 h2_{out} + b2 \right) \\
\end{align*}
We know $\frac{\partial h1_{out}}{\partial h1_{net}}$ and $\frac{\partial h1_{net}}{\partial w_1}$. \\ \\
Adding up everything,
\begin{equation*}
\frac{\partial E_T}{\partial w_1} = \frac{\partial E_{o1}}{\partial w_1} + \frac{\partial E_{o2}}{\partial w_2}
\end{equation*}
\begin{equation*}
\boxed{
	w_1 = w_1 - \eta\cdot \frac{\partial E_T}{\partial w_1}
} \qquad
\end{equation*}
Likewise, $\frac{\partial E_T}{\partial w_2}$, $\frac{\partial E_T}{\partial w_2}$, $\frac{\partial E_T}{\partial w_3}$, and $\frac{\partial E_T}{\partial w_4}$ can be computed.
\section{Discussion}
\begin{itemize}
\item $x$ is surface temperature (normalized between $0$ and $1$) and $1-x$ is the complement of that, together explaining the perfect discrimination between habitability classes as explained in our TSS above. The motivation of SBAF is derived from this fact of TSS. Using $kx^{\alpha}(1-x)^{1-\alpha}$ shall maximize the width of the two separating hyperplanes in the SVM used in TSS (See the proof below) as the kernel has a global maxima when $0 \leq \alpha \leq 1$. This is equivalent to the CDHS formulation when CD-HPF is written as $ y = kx^{\alpha}(1-x)^{\beta}$ where $\alpha+\beta=1, 0 \leq \alpha \leq 1, 0 \leq \beta \leq 1$, $k$ is suitably assumed to be $1$ (CRS condition), and the representation ensures global maxima (maximum width of the separating hyperplanes) under such constraints \citep{CDHPF2016,Potential}. 
\item The new activation function to be used for training a neural network for habitability classification boasts of an optima. Evidently, from the graphical simulations below, we observe less flattening of the function and therefore the formulation should be able to tackle local oscillations more easily as compared to the more generally used sigmoid function. Moreover, since $0 \leq \alpha \leq 1, 0 \leq x \leq 1, 0 \leq 1-x \leq 1$, the variable term in the denominator of SBAF, $kx^{\alpha}(1-x)^{1-\alpha}$ may be approximated to a first order polynomial. This may help us in circumventing expensive floating point operations without compromising the precision.
\item Need to show that the maxima is unique in the defined interval. This will circumvent the local maxima problem.
\end{itemize}
Habitability classification is a complex task. Even though the literature is replete with rich and sophisticated methods using both supervised \citep{Zighed2010} and unsupervised learning methods, the soft margin between classes, namely psychroplanet and mesoplanet makes the task of discrimination incredibly difficult. A sequence of recent explorations by Saha et. al. expanding previous work by Bora et. al. on using Machine Learning algorithm to construct and test planetary habitability functions with exoplanet data raises important questions. The 2018 paper (\citep{Potential}) analyzed the elasticity of the Cobb-Douglas Habitability Score (CDHS) and compared its performance with other machine learning algorithms. They demonstrated the robustness of their methods to identify potentially habitable planets \citep{saha2018machine} from exoplanet dataset. Given our little knowledge on exoplanets and habitability, these results and methods provide one important step toward automatically identifying objects of interest from large datasets by future ground and space observatories. The variable term in SBAF, $kx^{\alpha}(1-x)^{1-\alpha}$ is inspired from a history of modeling such terms as production functions and exploiting optimization principles in production economics, \citep{Saha2016}, \citep{Ginde2016}, \citep{ginde2015mining}. Complexities/bias in data may often necessitate devising classification methods to mitigate class imbalance, \citep{Mohanchandra2015} to improve upon the original method, \citep{vapnik1964}, \citep{Cortes1995} or manipulate confidence intervals \citep{Khaidem2016}. However, these improvisations led the authors to believe that, a general framework to train in forward and backward pass may turn out to be efficient. This is the primary reason to design a neural network with a novel activation function. We shall use the architecture to discriminate exoplanetary habitability \citep{schulze-makuch2018time}, \citep{SchulzeMakuch2011},
\citep{Irwin2014}, \citep{googlenasa}, \citep{hipparcosref}, \citep{phlref}. 

\section*{References}

\bibliography{mybibfile}

\begin{thebibliography}{25}
\expandafter\ifx\csname natexlab\endcsname\relax\def\natexlab#1{#1}\fi
\providecommand{\url}[1]{\texttt{#1}}
\providecommand{\href}[2]{#2}
\providecommand{\path}[1]{#1}
\providecommand{\DOIprefix}{doi:}
\providecommand{\ArXivprefix}{arXiv:}
\providecommand{\URLprefix}{URL: }
\providecommand{\Pubmedprefix}{pmid:}
\providecommand{\doi}[1]{\href{http://dx.doi.org/#1}{\path{#1}}}
\providecommand{\Pubmed}[1]{\href{pmid:#1}{\path{#1}}}
\providecommand{\bibinfo}[2]{#2}
\ifx\xfnm\relax \def\xfnm[#1]{\unskip,\space#1}\fi
\bibitem[{Agrawal et~al.(2018)Agrawal, Basak, Saha, Bora and
  Murthy}]{1804.11176}
\bibinfo{author}{Agrawal, S.}, \bibinfo{author}{Basak, S.},
  \bibinfo{author}{Saha, S.}, \bibinfo{author}{Bora, K.},
  \bibinfo{author}{Murthy, J.}, \bibinfo{year}{2018}.
\newblock \bibinfo{title}{A comparative analysis of the cobb-douglas
  habitability score (cdhs) with the earth similarity index (esi)}.
\newblock \href{http://arxiv.org/abs/arXiv:1804.11176}{\tt
  arXiv:arXiv:1804.11176}.
\bibitem[{Bains and Schulze-Makuch(2016)}]{Bains2016}
\bibinfo{author}{Bains, W.}, \bibinfo{author}{Schulze-Makuch, D.},
  \bibinfo{year}{2016}.
\newblock \bibinfo{title}{The cosmic zoo: The (near) inevitability of the
  evolution of complex, macroscopic life}.
\newblock \bibinfo{journal}{Life} \bibinfo{volume}{6}, \bibinfo{pages}{25}.
\newblock \URLprefix \url{https://doi.org/10.3390/life6030025},
  \DOIprefix\doi{10.3390/life6030025}.
\bibitem[{Basak et~al.(2018)Basak, Agrawal, Saha, Theophilus, Bora, Deshpande
  and Murthy}]{Insight}
\bibinfo{author}{Basak, S.}, \bibinfo{author}{Agrawal, S.},
  \bibinfo{author}{Saha, S.}, \bibinfo{author}{Theophilus, A.J.},
  \bibinfo{author}{Bora, K.}, \bibinfo{author}{Deshpande, G.},
  \bibinfo{author}{Murthy, J.}, \bibinfo{year}{2018}.
\newblock \bibinfo{title}{Saha-bora activation function: Habitability
  classification}.
\newblock \href{http://arxiv.org/abs/10.13140/RG.2.2.21081.62565}{\tt
  arXiv:10.13140/RG.2.2.21081.62565}.
\bibitem[{Bora et~al.(2016)Bora, Saha, Agrawal, Safonova, Routh and
  Narasimhamurthy}]{CDHPF2016}
\bibinfo{author}{Bora, K.}, \bibinfo{author}{Saha, S.},
  \bibinfo{author}{Agrawal, S.}, \bibinfo{author}{Safonova, M.},
  \bibinfo{author}{Routh, S.}, \bibinfo{author}{Narasimhamurthy, A.},
  \bibinfo{year}{2016}.
\newblock \bibinfo{title}{Cd-hpf: New habitability score via data analytic
  modeling}.
\newblock \bibinfo{journal}{Astronomy and Computing} \bibinfo{volume}{17},
  \bibinfo{pages}{129 -- 143}.
\newblock \URLprefix
  \url{http://www.sciencedirect.com/science/article/pii/S2213133716300865},
  \DOIprefix\doi{https://doi.org/10.1016/j.ascom.2016.08.001}.
\bibitem[{Cortes and Vapnik(1995)}]{Cortes1995}
\bibinfo{author}{Cortes, C.}, \bibinfo{author}{Vapnik, V.},
  \bibinfo{year}{1995}.
\newblock \bibinfo{title}{Support-vector networks}.
\newblock \bibinfo{journal}{Machine Learning} \bibinfo{volume}{20},
  \bibinfo{pages}{273--297}.
\newblock \URLprefix \url{https://doi.org/10.1023/A:1022627411411},
  \DOIprefix\doi{10.1023/A:1022627411411}.
\bibitem[{Elfwing et~al.(2018)Elfwing, Uchibe and Doya}]{ELFWING2018}
\bibinfo{author}{Elfwing, S.}, \bibinfo{author}{Uchibe, E.},
  \bibinfo{author}{Doya, K.}, \bibinfo{year}{2018}.
\newblock \bibinfo{title}{Sigmoid-weighted linear units for neural network
  function approximation in reinforcement learning}.
\newblock \bibinfo{journal}{Neural Networks} \URLprefix
  \url{http://www.sciencedirect.com/science/article/pii/S0893608017302976},
  \DOIprefix\doi{https://doi.org/10.1016/j.neunet.2017.12.012}.
\bibitem[{Ginde et~al.(2015)Ginde, Saha, Balasubramaniam, Harsha, Mathur,
  Dayasagar and Anand}]{ginde2015mining}
\bibinfo{author}{Ginde, G.}, \bibinfo{author}{Saha, S.},
  \bibinfo{author}{Balasubramaniam, C.}, \bibinfo{author}{Harsha, R.},
  \bibinfo{author}{Mathur, A.}, \bibinfo{author}{Dayasagar, B.},
  \bibinfo{author}{Anand, M.}, \bibinfo{year}{2015}.
\newblock \bibinfo{title}{Mining massive databases for computation of
  scholastic indices: Model and quantify internationality and influence
  diffusion of peer-reviewed journals}, in: \bibinfo{booktitle}{Proceedings of
  the fourth national conference of Institute of Scientometrics, SIoT}.
\bibitem[{Ginde et~al.(2016)Ginde, Saha, Mathur, Venkatagiri, Vadakkepat,
  Narasimhamurthy and Daya~Sagar}]{Ginde2016}
\bibinfo{author}{Ginde, G.}, \bibinfo{author}{Saha, S.},
  \bibinfo{author}{Mathur, A.}, \bibinfo{author}{Venkatagiri, S.},
  \bibinfo{author}{Vadakkepat, S.}, \bibinfo{author}{Narasimhamurthy, A.},
  \bibinfo{author}{Daya~Sagar, B.S.}, \bibinfo{year}{2016}.
\newblock \bibinfo{title}{Scientobase: a framework and model for computing
  scholastic indicators of non-local influence of journals via native data
  acquisition algorithms}.
\newblock \bibinfo{journal}{Scientometrics} \bibinfo{volume}{108},
  \bibinfo{pages}{1479--1529}.
\newblock \URLprefix \url{https://doi.org/10.1007/s11192-016-2006-2},
  \DOIprefix\doi{10.1007/s11192-016-2006-2}.
\bibitem[{Irwin et~al.(2014)Irwin, M{\'{e}}ndez, Fair{\'{e}}n and
  Schulze-Makuch}]{Irwin2014}
\bibinfo{author}{Irwin, L.}, \bibinfo{author}{M{\'{e}}ndez, A.},
  \bibinfo{author}{Fair{\'{e}}n, A.}, \bibinfo{author}{Schulze-Makuch, D.},
  \bibinfo{year}{2014}.
\newblock \bibinfo{title}{Assessing the possibility of biological complexity on
  other worlds, with an estimate of the occurrence of complex life in the milky
  way galaxy}.
\newblock \bibinfo{journal}{Challenges} \bibinfo{volume}{5},
  \bibinfo{pages}{159--174}.
\newblock \URLprefix \url{https://doi.org/10.3390/challe5010159},
  \DOIprefix\doi{10.3390/challe5010159}.
\bibitem[{Khaidem et~al.(2016)Khaidem, Saha, Basak and Dey}]{Khaidem2016}
\bibinfo{author}{Khaidem, L.}, \bibinfo{author}{Saha, S.},
  \bibinfo{author}{Basak, S.}, \bibinfo{author}{Dey, S.R.},
  \bibinfo{year}{2016}.
\newblock \bibinfo{title}{Predicting the direction of stock market prices using
  random forest}.
\newblock \URLprefix
  \url{"https://www.researchgate.net/publication/301818771_Predicting_the_direction_of_stock_market_prices_using_random_forest"}.
\bibitem[{Lippmann(1994)}]{neural}
\bibinfo{author}{Lippmann, R.}, \bibinfo{year}{1994}.
\newblock \bibinfo{title}{Book review: "neural networks, a comprehensive
  foundation", by simon haykin}.
\newblock \bibinfo{journal}{International Journal of Neural Systems}
  \bibinfo{volume}{05}, \bibinfo{pages}{363--364}.
\newblock \URLprefix \url{https://doi.org/10.1142/s0129065794000372},
  \DOIprefix\doi{10.1142/s0129065794000372}.
\bibitem[{M{\'{e}}ndez(2011)}]{hipparcosref}
\bibinfo{author}{M{\'{e}}ndez, A.}, \bibinfo{year}{2011}.
\newblock \bibinfo{title}{The night sky of exoplanets} \URLprefix
  \url{http://phl.upr.edu/library/notes/syntheticstars}.
\bibitem[{M{\'{e}}ndez(2018)}]{phlref}
\bibinfo{author}{M{\'{e}}ndez, A.}, \bibinfo{year}{2018}.
\newblock \bibinfo{title}{The habitable exoplanets catalog} \URLprefix
  \url{http://phl.upr.edu/hec}.
\bibitem[{Mohanchandra et~al.(2015)Mohanchandra, Saha, Murthy and
  Lingaraju}]{Mohanchandra2015}
\bibinfo{author}{Mohanchandra, K.}, \bibinfo{author}{Saha, S.},
  \bibinfo{author}{Murthy, K.S.}, \bibinfo{author}{Lingaraju, G.},
  \bibinfo{year}{2015}.
\newblock \bibinfo{title}{Distinct adoption of k-nearest neighbour and support
  vector machine in classifying {EEG} signals of mental tasks}.
\newblock \bibinfo{journal}{International Journal of Intelligent Engineering
  Informatics} \bibinfo{volume}{3}, \bibinfo{pages}{313}.
\newblock \URLprefix \url{https://doi.org/10.1504/ijiei.2015.073064},
  \DOIprefix\doi{10.1504/ijiei.2015.073064}.
\bibitem[{{Saha} et~al.(2017){Saha}, {Basak}, {Bora}, {Safonova}, {Agrawal},
  {Sarkar} and {Murthy}}]{Potential}
\bibinfo{author}{{Saha}, S.}, \bibinfo{author}{{Basak}, S.},
  \bibinfo{author}{{Bora}, K.}, \bibinfo{author}{{Safonova}, M.},
  \bibinfo{author}{{Agrawal}, S.}, \bibinfo{author}{{Sarkar}, P.},
  \bibinfo{author}{{Murthy}, J.}, \bibinfo{year}{2017}.
\newblock \bibinfo{title}{{Theoretical Validation of Potential Habitability via
  Analytical and Boosted Tree Methods: An Optimistic Study on Recently
  Discovered Exoplanets}}.
\newblock \bibinfo{journal}{ArXiv e-prints}
  \href{http://arxiv.org/abs/1712.01040}{\tt arXiv:1712.01040}.
\bibitem[{Saha et~al.(2018a)Saha, Bora, Basak, Mathur and Agrawal}]{sbaf}
\bibinfo{author}{Saha, S.}, \bibinfo{author}{Bora, K.}, \bibinfo{author}{Basak,
  S.}, \bibinfo{author}{Mathur, A.}, \bibinfo{author}{Agrawal, S.},
  \bibinfo{year}{2018}a.
\newblock \bibinfo{title}{Habitability classification of exoplanets: A machine
  learning insight}.
\newblock \href{http://arxiv.org/abs/arXiv:1805.08810}{\tt
  arXiv:arXiv:1805.08810}.
\bibitem[{Saha et~al.(2018b)Saha, Bora, Basak, Srinivasa, Safonova, Murthy and
  Agrawal}]{saha2018machine}
\bibinfo{author}{Saha, S.}, \bibinfo{author}{Bora, K.}, \bibinfo{author}{Basak,
  S.}, \bibinfo{author}{Srinivasa, G.}, \bibinfo{author}{Safonova, M.},
  \bibinfo{author}{Murthy, J.}, \bibinfo{author}{Agrawal, S.},
  \bibinfo{year}{2018}b.
\newblock \bibinfo{title}{Ebook-astroinformatics series machine learning in
  astronomy: A workman's manual}.
\newblock \URLprefix
  \url{"https://www.researchgate.net/publication/322926268_EBOOK-ASTROINFORMATICS_SERIES_MACHINE_LEARNING_IN_ASTRONOMY_A_WORKMAN'S_MANUAL"}.
\bibitem[{Saha et~al.(2016)Saha, Sarkar, Dwivedi, Dwivedi, Narasimhamurthy and
  Roy}]{Saha2016}
\bibinfo{author}{Saha, S.}, \bibinfo{author}{Sarkar, J.},
  \bibinfo{author}{Dwivedi, A.}, \bibinfo{author}{Dwivedi, N.},
  \bibinfo{author}{Narasimhamurthy, A.M.}, \bibinfo{author}{Roy, R.},
  \bibinfo{year}{2016}.
\newblock \bibinfo{title}{A novel revenue optimization model to address the
  operation and maintenance cost of a data center}.
\newblock \bibinfo{journal}{Journal of Cloud Computing} \bibinfo{volume}{5},
  \bibinfo{pages}{1}.
\newblock \URLprefix \url{https://doi.org/10.1186/s13677-015-0050-8},
  \DOIprefix\doi{10.1186/s13677-015-0050-8}.
\bibitem[{Saha et~al.(2018c)Saha, Sarkar, Mathur and Basak}]{1803.04644}
\bibinfo{author}{Saha, S.}, \bibinfo{author}{Sarkar, P.},
  \bibinfo{author}{Mathur, A.}, \bibinfo{author}{Basak, S.},
  \bibinfo{year}{2018}c.
\newblock \bibinfo{title}{Model visualization in understanding rapid growth of
  a journal in an emerging area}.
\newblock \href{http://arxiv.org/abs/arXiv:1803.04644}{\tt
  arXiv:arXiv:1803.04644}.
\bibitem[{Schulze-Makuch and Bains(2018)}]{schulze-makuch2018time}
\bibinfo{author}{Schulze-Makuch, D.}, \bibinfo{author}{Bains, W.},
  \bibinfo{year}{2018}.
\newblock \bibinfo{title}{Time to consider search strategies for complex life
  on exoplanets}.
\newblock \bibinfo{journal}{Nature Astronomy} , \bibinfo{pages}{1--2}\URLprefix
  \url{http:https://doi.org/10.1038/s41550-018-0476-2},
  \DOIprefix\doi{10.1038/s41550-018-0476-2}.
\bibitem[{Schulze-Makuch et~al.(2011)Schulze-Makuch, M{\'{e}}ndez,
  Fair{\'{e}}n, von Paris, Turse, Boyer, Davila, de~Sousa~Ant{\'{o}}nio,
  Catling and Irwin}]{SchulzeMakuch2011}
\bibinfo{author}{Schulze-Makuch, D.}, \bibinfo{author}{M{\'{e}}ndez, A.},
  \bibinfo{author}{Fair{\'{e}}n, A.G.}, \bibinfo{author}{von Paris, P.},
  \bibinfo{author}{Turse, C.}, \bibinfo{author}{Boyer, G.},
  \bibinfo{author}{Davila, A.F.}, \bibinfo{author}{de~Sousa~Ant{\'{o}}nio,
  M.R.}, \bibinfo{author}{Catling, D.}, \bibinfo{author}{Irwin, L.N.},
  \bibinfo{year}{2011}.
\newblock \bibinfo{title}{A two-tiered approach to assessing the habitability
  of exoplanets}.
\newblock \bibinfo{journal}{Astrobiology} \bibinfo{volume}{11},
  \bibinfo{pages}{1041--1052}.
\newblock \URLprefix \url{https://doi.org/10.1089/ast.2010.0592},
  \DOIprefix\doi{10.1089/ast.2010.0592}.
\bibitem[{Shallue and Vanderburg(2018)}]{googlenasa}
\bibinfo{author}{Shallue, C.J.}, \bibinfo{author}{Vanderburg, A.},
  \bibinfo{year}{2018}.
\newblock \bibinfo{title}{Identifying exoplanets with deep learning: A
  five-planet resonant chain around kepler-80 and an eighth planet around
  kepler-90}.
\newblock \bibinfo{journal}{The Astronomical Journal} \bibinfo{volume}{155},
  \bibinfo{pages}{94}.
\newblock \URLprefix \url{http://stacks.iop.org/1538-3881/155/i=2/a=94}.
\bibitem[{Vapnik and Chervonenkis(1964)}]{vapnik1964}
\bibinfo{author}{Vapnik, V.N.}, \bibinfo{author}{Chervonenkis, A.Y.},
  \bibinfo{year}{1964}.
\newblock \bibinfo{title}{On a class of perceptrons}.
\newblock \bibinfo{journal}{Automation and Remote Control} \bibinfo{volume}{1},
  \bibinfo{pages}{103--109}.
\bibitem[{Younger et~al.()Younger, Hochreiter and Conwell}]{back}
\bibinfo{author}{Younger, A.}, \bibinfo{author}{Hochreiter, S.},
  \bibinfo{author}{Conwell, P.}, .
\newblock \bibinfo{title}{Meta-learning with backpropagation}, in:
  \bibinfo{booktitle}{{IJCNN} 01. International Joint Conference on Neural
  Networks. Proceedings (Cat. No.01CH37222)}, \bibinfo{publisher}{{IEEE}}.
\newblock \URLprefix \url{https://doi.org/10.1109/ijcnn.2001.938471},
  \DOIprefix\doi{10.1109/ijcnn.2001.938471}.
\bibitem[{Zighed et~al.(2010)Zighed, Ritschard and Marcellin}]{Zighed2010}
\bibinfo{author}{Zighed, D.A.}, \bibinfo{author}{Ritschard, G.},
  \bibinfo{author}{Marcellin, S.}, \bibinfo{year}{2010}.
\newblock \bibinfo{title}{Asymmetric and Sample Size Sensitive Entropy Measures
  for Supervised Learning}. \bibinfo{publisher}{Springer Berlin Heidelberg},
  \bibinfo{address}{Berlin, Heidelberg}.
\newblock pp. \bibinfo{pages}{27--42}.
\newblock \URLprefix \url{https://doi.org/10.1007/978-3-642-05183-8_2},
  \DOIprefix\doi{10.1007/978-3-642-05183-8_2}.

\end{thebibliography}
\end{document}